# Example-Based Optimization of Surface-Generation Tables

Christer Samuelsson
*Universität des Saarlandes*

## Abstract

A method is given that "inverts" a logic grammar and displays it from the point of view of the logical form, rather than from that of the word string. LR-compiling techniques are used to allow a recursive-descent generation algorithm to perform "functor merging" much in the same way as an LR parser performs prefix merging. This is an improvement on the semantic-head-driven generator that results in a much smaller search space. The amount of semantic lookahead can be varied, and appropriate tradeoff points between table size and resulting nondeterminism can be found automatically. This can be done by removing all spurious nondeterminism for input sufficiently close to the examples of a training corpus, and large portions of it for other input, while preserving completeness. [1]

## 1  Introduction

With the emergence of fast algorithms and optimization techniques for syntactic analysis, such as the use of explanation-based learning in conjunction with LR parsing, see (Samuelsson & Rayner 1991) and subsequent work, surface generation has become a major bottleneck in NLP systems. Surface generation will here be viewed as the inverse problem of syntactic analysis and subsequent semantic interpretation. The latter consists in constructing some semantic representation of an input word-string based on the syntactic and semantic rules of a formal grammar. In this article, we will limit ourselves to logic grammars that attribute word strings with expressions in some logical formalism represented as terms with a functor-argument structure. The surface generation problem then consists in assigning an output

---

[1] I wish to thank greatly Gregor Erbach, Jussi Karlgren, Manny Rayner, Hans Uszkoreit, Mats Wirén and the anonymous reviewers of ACL, EACL, IJCAI and RANLP for valuable feedback on previous versions of this article. Special credit is due to Kristina Striegnitz, who assisted with the implementation.

Parts of this article have previously appeared as (Samuelsson 1995). The presented work was funded by the N3 "Bidirektionale Linguistische Deduktion (BiLD)" project in the Sonderforschungsbereich 314 *Künstliche Intelligenz — Wissensbasierte Systeme*.

word-string to such a term. This is a common scenario in conjunction with for example transfer-based machine-translation systems employing reversible grammars, and it is different from that when a deep generator or a text planner is available to guide the surface generator. In general, both these mappings are many-to-many: a word string that can be mapped to several distinct logical forms is said to be ambiguous. A logical form that can be assigned to several different word strings is said to have multiple paraphrases.

We want to create a generation algorithm that generates a word string by recursively descending through a logical form, while delaying the choice of grammar rules to apply as long as possible. This means that we want to process different rules or rule combinations that introduce the same piece of semantics in parallel until they branch apart. This will reduce the amount of spurious search, since we will gain more information about the rest of the logical form before having to commit to a particular grammar rule. In practice, this means that we want to perform "functor merging" much in the same ways as an LR parser performs prefix merging by employing parsing tables compiled from the grammar. One obvious way of doing this is to use LR-compilation techniques to compile generation tables. This will however require that we reformulate the grammar from the point of view of the logical form, rather than from that of the word string from which it is normally displayed.

The rest of the paper is structured as follows: We will first review basic LR compilation of parsing tables in Section 2. The grammar-inversion procedure turns out to be most easily explained in terms of the semantic-head-driven generation (SHDG) algorithm. We will therefore proceed to outline the SHDG algorithm in Section 3. The grammar inversion itself is described in Section 4, while LR compilation of generation tables is discussed in Section 5. The generation algorithm is presented in Section 6. The example-based optimization technique turns out to be most easily explained as a straight-forward extension of a simpler optimization technique predating it, why this simpler technique is given in Section 7. This extension is described in Section 8 and the relation between this example-based optimization technique and explanation-based learning is discussed in Section 9.

## 2   LR Compilation for Parsing

LR compilation in general is well-described in for example (Aho et al. 1986:215–247). Here we will only sketch out the main ideas.

An LR parser is basically a pushdown automaton, i.e., it has a pushdown stack in addition to a finite set of internal states and a reader head for scanning the input string from left to right one symbol at a time. The stack is used in a characteristic way: The items on the stack consist of alternating grammar symbols and states. The current state is simply the state on top of the stack. The most distinguishing feature of an LR parser is however the form of the transition relation — the action and goto tables. A nondeterministic LR parser can in each step perform one of four basic actions. In state $S$ with lookahead symbol[2] $Sym$ it can:

1. **accept**: Halt and signal success.

2. **error**: Fail and backtrack.

3. **shift** $S_2$: Consume the input symbol $Sym$, push it onto the stack, and transit to state $S_2$ by pushing it onto the stack.

4. **reduce** $R$: Pop off two items from the stack for each grammar symbol in the RHS of grammar rule $R$, inspect the stack for the old state $S_1$ now on top of the stack, push the $LHS$ of rule $R$ onto the stack, and transit to state $S_2$ determined by $goto(S_1,LHS,S_2)$ by pushing $S_2$ onto the stack.

Consider the small sample grammar given in Figure 1. To make this simple grammar slightly more interesting, the recursive Rule 1, $S \rightarrow S\ QM$, allows the addition of a question mark ($QM$) to the end of a sentence ($S$), as in *John sleeps?*. The LHS $S$ is then interpreted as a yes-no question version of the RHS $S$.

Each internal state consists of a set of dotted items. Each item in turn corresponds to a grammar rule. The current string position is indicated by a dot. For example, Rule 2, $S \rightarrow NP\ VP$, yields the item $S \Rightarrow NP \bullet VP$, which corresponds to just having found an $NP$ and now searching for a $VP$.

In the compilation phase, new states are induced from old ones: For the indicated string position, a possible grammar symbol is selected and the dot is advanced one step in all items where this particular grammar symbol immediately follows the dot, and the resulting new items will constitute the kernel of the new state. Non-kernel items are added to these by selecting grammar rules whose LHS match grammar symbols at the new string position in the new items. In each non-kernel item, the dot is at the beginning of the rule. If a set of items is constructed that already exists, then this search branch is abandoned and the recursion terminates.

---

[2]The lookahead symbol is the next symbol in the input string, i.e., the symbol under the reader head.

| | | | | | | |
|---|---|---|---|---|---|---|
| $S$ | $\rightarrow$ | $S\ QM$ | $1$ | $NP$ | $\rightarrow$ | $John$ |
| $S$ | $\rightarrow$ | $NP\ VP$ | $2$ | $NP$ | $\rightarrow$ | $Mary$ |
| $VP$ | $\rightarrow$ | $VP\ PP$ | $3$ | $NP$ | $\rightarrow$ | $Paris$ |
| $VP$ | $\rightarrow$ | $VP\ AdvP$ | $4$ | $V_i$ | $\rightarrow$ | $sleeps$ |
| $VP$ | $\rightarrow$ | $V_i$ | $5$ | $V_t$ | $\rightarrow$ | $sees$ |
| $VP$ | $\rightarrow$ | $V_t\ NP$ | $6$ | $P$ | $\rightarrow$ | $in$ |
| $PP$ | $\rightarrow$ | $P\ NP$ | $7$ | $AdvP$ | $\rightarrow$ | $today$ |
| | | | | $QM$ | $\rightarrow$ | $?$ |

Fig. 1. *Sample grammar*

State 1
$S' \Rightarrow \bullet\ S$
$S \Rightarrow \bullet\ S\ QM$
$S \Rightarrow \bullet\ NP\ VP$

State 2
$S' \Rightarrow S\ \bullet$
$S \Rightarrow S\ \bullet\ QM$

State 3
$S \Rightarrow NP\ \bullet\ VP$
$VP \Rightarrow \bullet\ VP\ PP$
$VP \Rightarrow \bullet\ VP\ AdvP$
$VP \Rightarrow \bullet\ V_i$
$VP \Rightarrow \bullet\ V_t\ NP$

State 4
$S \Rightarrow NP\ VP\ \bullet$
$VP \Rightarrow VP\ \bullet\ PP$
$VP \Rightarrow VP\ \bullet\ AdvP$
$PP \Rightarrow \bullet\ P\ NP$

State 5
$VP \Rightarrow VP\ PP\ \bullet$

State 6
$VP \Rightarrow VP\ AdvP\ \bullet$

State 7
$PP \Rightarrow P\ \bullet\ NP$

State 8
$PP \Rightarrow P\ NP\ \bullet$

State 9
$VP \Rightarrow V_i\ \bullet$

State 10
$VP \Rightarrow V_t\ \bullet\ NP$

State 11
$VP \Rightarrow V_t\ NP\ \bullet$

State 12
$S \Rightarrow S\ QM\ \bullet$

Fig. 2. *LR-parsing states for the sample grammar*

The state-construction phase starts off by creating an initial set consisting of a single dummy kernel item and its non-kernel closure. This is State 1 in Figure 2. The dummy item introduces a dummy top grammar symbol as its LHS, while the RHS consists of the old top symbol, and the dot is at the beginning of the rule. In the example, this is the item $S' \Rightarrow \bullet \, S$. The rest of the states are induced from the initial state. The states resulting from the sample grammar of Figure 1 are shown in Figure 2, and these in turn will yield the parsing tables of Figure 3. The entry "s3" in the action table, for example, should be interpreted as "shift the lookahead symbol onto the stack and transit to State 3". The entry "r7" should be interpreted as "reduce by Rule 7". The accept action is denoted "acc". The goto entries, like "g4", simply indicate what state to transit to once a nonterminal of that type has been constructed.

| | NP | VP | PP | AdvP | $V_i$ | $V_t$ | P | S | QM | eos |
|---|---|---|---|---|---|---|---|---|---|---|
| 1 | s3 | | | | | | | g2 | | |
| 2 | | | | | | | | | s12 | acc |
| 3 | | g4 | | | s9 | s10 | | | | |
| 4 | | | g5 | s6 | | | s7 | | r2 | r2 |
| 5 | | | | r3 | | | r3 | | r3 | r3 |
| 6 | | | | r4 | | | r4 | | r4 | r4 |
| 7 | s8 | | | | | | | | | |
| 8 | | | | r7 | | | r7 | | r7 | r7 |
| 9 | | | | r5 | | | r5 | | r5 | r5 |
| 10 | s11 | | | | | | | | | |
| 11 | | | | r6 | | | r6 | | r6 | r6 |
| 12 | | | | | | | | | r1 | r1 |

Fig. 3. *LR-parsing tables for the sample grammar*

In conjunction with grammar formalisms employing complex feature structures, this procedure is associated with a number of interesting problems, many of which are discussed in (Nakazawa 1991) and (Samuelsson 1994c). For example, the termination criterion must be modified: If a new set of items is constructed that is *more specific* than an existing one, then this search branch is abandoned and the recursion terminates. If, on the other hand, it is *more general*, then it replaces the old one.

## 3   The Semantic-Head-Driven Generation Algorithm

Generators found in large-scale systems such as the DFKI DISCO system (Uszkoreit *et al* 1994), or the SRI Core Language Engine (Alshawi (ed.) 1992:268–275), tend typically to be based on the semantic-head-driven generation (SHDG) algorithm. The SHDG algorithm is well-described in (Shieber *et al* 1990); here we will only outline the main features.

The grammar rules of Figure 1 have been attributed with logical forms as shown in Figure 4. The notation has been changed so that each constituent consists of a quadruple $\langle Cat, \mathtt{Sem}, W_0, W_1 \rangle$, where $W_0$ and $W_1$ form a difference list representing the word string that $Cat$ spans, and $\mathtt{Sem}$ is the logical form. For example, the logical form corresponding to the LHS $S$ of the $\langle S, \mathtt{mod(X,Y)}, W_0, W \rangle \rightarrow \langle S, \mathtt{X}, W_0, W_1 \rangle \langle QM, \mathtt{Y}, W_1, W \rangle$ rule, consists of a modifier $\mathtt{Y}$ added to the logical form $\mathtt{X}$ of the RHS $S$. As we can see from the last grammar rule, this modifier is in turn realized as $\mathtt{ynq}$.

---

$$\langle S, \mathtt{mod(X,Y)}, W_0, W \rangle \;\rightarrow\; \langle S, \mathtt{X}, W_0, W_1 \rangle \; \langle QM, \mathtt{Y}, W_1, W \rangle \qquad 1$$

$$\langle S, \mathtt{Y}, W_0, W \rangle \;\rightarrow\; \langle NP, \mathtt{X}, W_0, W_1 \rangle \; \langle VP, \mathtt{X\hat{}Y}, W_1, W \rangle \qquad 2$$

$$\langle VP, \mathtt{X\hat{}mod(Y,Z)}, W_0, W \rangle \;\rightarrow\; \langle VP, \mathtt{X\hat{}Y}, W_0, W_1 \rangle \; \langle AdvP, \mathtt{Z}, W_1, W \rangle \qquad 3$$

$$\langle VP, \mathtt{X\hat{}mod(Y,Z)}, W_0, W \rangle \;\rightarrow\; \langle VP, \mathtt{X\hat{}Y}, W_0, W_1 \rangle \; \langle PP, \mathtt{Z}, W_1, W \rangle \qquad 4$$

$$\langle VP, \mathtt{X}, W_0, W \rangle \;\rightarrow\; \langle V_i, \mathtt{X}, W_0, W \rangle \qquad 5$$

$$\langle VP, \mathtt{Y}, W_0, W \rangle \;\rightarrow\; \langle V_t, \mathtt{X\hat{}Y}, W_0, W_1 \rangle \; \langle NP, \mathtt{X}, W_1, W \rangle \qquad 6$$

$$\langle PP, \mathtt{Y}, W_0, W \rangle \;\rightarrow\; \langle P, \mathtt{X\hat{}Y}, W_0, W_1 \rangle \; \langle NP, \mathtt{X}, W_1, W \rangle \qquad 7$$

$$\langle NP, \mathtt{john}, [John|W], W \rangle \;\rightarrow\; John$$

$$\langle NP, \mathtt{mary}, [Mary|W], W \rangle \;\rightarrow\; Mary$$

$$\langle NP, \mathtt{paris}, [Paris|W], W \rangle \;\rightarrow\; Paris$$

$$\langle V_i, \mathtt{X\hat{}sleep(X)}, [sleeps|W], W \rangle \;\rightarrow\; sleeps$$

$$\langle V_t, \mathtt{X\hat{}Y\hat{}see(X,Y)}, [see|W], W \rangle \;\rightarrow\; sees$$

$$\langle P, \mathtt{X\hat{}in(X)}, [in|W], W \rangle \;\rightarrow\; in$$

$$\langle AdvP, \mathtt{today}, [today|W], W \rangle \;\rightarrow\; today$$

$$\langle QM, \mathtt{ynq}, [?|W], W \rangle \;\rightarrow\; ?$$

Fig. 4. *Sample grammar with semantics*

---

For the SHDG algorithm, the grammar is divided into chain rules and non-chain rules: Chain rules have a distinguished RHS constituent, the semantic head, that has the same logical form as the LHS constituent, modulo $\lambda$-abstractions; non-chain rules lack such a constituent. In particular, lexicon entries are non-chain rules, since they do not have any RHS constituents at all. This distinction is made since the generation algorithm treats the two rule types quite differently. In the example grammar, rules 2 and 5 through 7 are chain rules, while the remaining ones are non-chain rules.

A simple semantic-head-driven generator might work as follows: Given a grammar symbol and a piece of logical form, the generator looks for a non-chain rule with the given semantics. The constituents of the RHS of that rule are then generated recursively, after which the LHS is connected to the given grammar symbol using chain rules. At each application of a chain rule, the rest of the RHS constituents, i.e., the non-head constituents, are generated recursively. The particular combination of connecting chain rules used is often referred to as a chain. The generator starts off with the top symbol of the grammar and the logical form corresponding to the string that is to be generated.

The inherent problem with the SHDG algorithm is that each rule combination is tried in turn, while the possibilities of prefiltering are rather limited, leading to a large amount of spurious search. The generation algorithm presented in the current article does not suffer from this problem; what the new algorithm in effect does is to process all chains from a particular set of grammar symbols down to some particular piece of logical form in parallel before any rule is applied, rather than to construct and try each one separately in turn.

## 4    Grammar Inversion

Before we can invert the grammar, we must put it in normal form. We will use a variant of chain and non-chain rules, namely functor-introducing rules corresponding to non-chain rules, and argument-filling rules corresponding to chain rules. The inversion step is based on the assumption that there are no other types of rules.

Since the generator will work by recursive descent through the logical form, we wish to rearrange the grammar so that arguments are generated together with their functors. To this end we introduce another difference list $A_0$ and $A$ to pass down the arguments introduced by argument-filling rules to the corresponding functor-introducing rules. Here the latter rules are assumed to be lexical, following the tradition in GPSG where the presence of the SUBCAT feature implies a preterminal grammar symbol, see e.g. (Gazdar et al. 1985:33), but this is really immaterial for the algorithm.

The grammar of Figure 4 is shown in normal form in Figure 5. The grammar is compiled into this form by inspecting the flow of arguments

**Functor-introducing rules**

$\langle S, \mathtt{mod(X,Y)}, W_0, W, \epsilon, \epsilon \rangle \;\rightarrow\; \langle S, \mathtt{X}, W_0, W_1, \epsilon, \epsilon \rangle \; \langle QM, \mathtt{Y}, W_1, W, \epsilon, \epsilon \rangle$   1

$\langle VP, \mathtt{X\char`^mod(Y,Z)}, W_0, W, A_0, A \rangle \;\rightarrow$   3
       $\langle VP, \mathtt{X\char`^Y}, W_0, W_1, A_0, A \rangle \; \langle AdvP, \mathtt{Z}, W_1, W, \epsilon, \epsilon \rangle$

$\langle VP, \mathtt{X\char`^mod(Y,Z)}, W_0, W, A_0, A \rangle \;\rightarrow$   4
       $\langle VP, \mathtt{X\char`^Y}, W_0, W_1, A_0, A \rangle \; \langle PP, \mathtt{Z}, W_1, W, \epsilon, \epsilon \rangle$

$\langle NP, \mathtt{john}, [John|W], W, A, \epsilon \rangle \;\rightarrow\; A$

$\langle NP, \mathtt{mary}, [Mary|W], W, A, \epsilon \rangle \;\rightarrow\; A$

$\langle NP, \mathtt{paris}, [Paris|W], W, A, \epsilon \rangle \;\rightarrow\; A$

$\langle V_i, \mathtt{X\char`^sleep(X)}, [sleeps|W], W, A, \epsilon \rangle \;\rightarrow\; A$

$\langle V_t, \mathtt{X\char`^Y\char`^see(X,Y)}, [see|W], W, A, \epsilon \rangle \;\rightarrow\; A$

$\langle P, \mathtt{X\char`^in(X)}, [in|W], W, A, \epsilon \rangle \;\rightarrow\; A$

$\langle AdvP, \mathtt{today}, [today|W], W, A, \epsilon \rangle \;\rightarrow\; A$

$\langle QM, \mathtt{ynq}, [?|W], W, A, \epsilon \rangle \;\rightarrow\; A$

**Argument-filling rules**

$\langle S, \mathtt{Y}, W_0, W, \epsilon, \epsilon \rangle \;\rightarrow\; \langle VP, \mathtt{X\char`^Y}, W_1, W, [\langle NP, \mathtt{X}, W_0, W_1 \rangle], \epsilon \rangle$   2

$\langle VP, \mathtt{X}, W_0, W, A_0, A \rangle \;\rightarrow\; \langle V_i, \mathtt{X}, W_0, W, A_0, A \rangle$   5

$\langle VP, \mathtt{Y}, W_0, W, A_0, A \rangle \;\rightarrow\; \langle V_t, \mathtt{X\char`^Y}, W_0, W_1, [\langle NP, \mathtt{X}, W_1, W \rangle | A_0], A \rangle$   6

$\langle PP, \mathtt{Y}, W_0, W, A_0, A \rangle \;\rightarrow\; \langle P, \mathtt{X\char`^Y}, W_0, W_1, [\langle NP, \mathtt{X}, W_1, W \rangle | A_0], A \rangle$   7

Fig. 5. *Sample grammar in normal form*

---

through the logical forms of the constituents of each rule. In the functor-introducing rules, the RHS is rearranged to mirror the argument order of the LHS logical form. The argument-filling rules have only one RHS constituent — the semantic head — and the rest of the original RHS constituents are added to the argument list of the head constituent. Note, for example, how the *NP* is added to the argument list of the *VP* in Rule 2, or to the argument list of the *P* in Rule 7. This is done automatically, although currently, the exact flow of arguments is specified manually.

We assume that there are no purely argument-filling cycles. For rules that actually fill in arguments, this is obviously impossible, since the number of arguments decreases strictly. For the slightly degenerate case of argument-filling rules which only pass along the logical form, such as the $\langle VP, \mathtt{X} \rangle \rightarrow \langle V_i, \mathtt{X} \rangle$ rule, this is equivalent to the off-line parsability requirement, (Kaplan & Bresnan 1982:264–266).[3] We require this in order to avoid an infinite number of chains, since each possible chain will be expanded out in the inversion step. Since subcategorization lists of verbs are bounded in length, PATR II style *VP* rules do not pose a serious problem, which on the other hand the "adjunct-as-argument" approach taken in (Bouma & van Noord 1994) may do. However, this problem is common to a number of other generation algorithms, including the SHDG algorithm.

---

[3] If the RHS $V_i$ were a *VP*, we would have a purely argument-filling cycle of length 1.

Let us return to the scenario for the SHDG algorithm given at the end of Section 3: We have a piece of logical form and a grammar symbol, and we wish to connect a non-chain rule with this particular logical form to the given grammar symbol through a chain. We will generalize this scenario just slightly to the case where a set of grammar symbols is given, rather than a single one.

Each inverted rule will correspond to a particular chain of argument-filling (chain) rules connecting a functor-introducing (non-chain) rule introducing this logical form to a grammar symbol in the given set. The arguments introduced by this chain will be collected and passed down to the functors that consume them in order to ensure that each of the inverted rules has a RHS matching the structure of the LHS logical form. The normalized sample grammar of Figure 5 will result in the inverted grammar of Figure 6. Note how the right-hand sides reflect the argument structure of the left-hand-side logical forms. As mentioned previously, the collected arguments are currently assumed to correspond to functors introduced by lexical entries, but the procedure can readily be modified to accommodate grammar rules with a non-empty RHS, where some of the arguments are consumed by the LHS logical form.

The grammar inversion step is combined with the LR-compilation step. This is convenient for several reasons: Firstly, the termination criteria and the database maintenance issues are the same in both steps. Secondly, since the LR-compilation step employs a top-down rule-invocation scheme, this will ensure that the arguments are passed down to the corresponding functors. In fact, invoking inverted grammar rules merely requires first invoking a chain of argument-filling rules and then terminating it with a functor-introducing rule.

## 5   LR Compilation for Generation

Just as when compiling LR-parsing tables, the compiler operates on sets of dotted items. Each item consists of a partially processed inverted grammar rule, with a dot marking the current position. Here the current position is

$\langle S, \mathtt{mod(X,Y)}, W_0, W, \epsilon, \epsilon \rangle \rightarrow$
$\quad \langle S, \mathtt{X}, W_0, W_1, \epsilon, \epsilon \rangle \ \langle QM, \mathtt{Y}, W_1, W, \epsilon, \epsilon \rangle$

$\langle S, \mathtt{mod(Y,Z)}, W_0, W, \epsilon, \epsilon \rangle \rightarrow$
$\quad \langle VP, \mathtt{X\hat{}Y}, W_1, W_2, [\langle NP, \mathtt{X}, W_0, W_1 \rangle], \epsilon \rangle \ \langle AdvP, \mathtt{Z}, W_2, W, \epsilon, \epsilon \rangle$

$\langle S, \mathtt{mod(Y,Z)}, W_0, W, \epsilon, \epsilon \rangle \rightarrow$
$\quad \langle VP, \mathtt{X\hat{}Y}, W_1, W_2, [\langle NP, \mathtt{X}, W_0, W_1 \rangle], \epsilon \rangle \ \langle PP, \mathtt{Z}, W_2, W, \epsilon, \epsilon \rangle$

$\langle VP, \mathtt{X\hat{}mod(Y,Z)}, W_1, W, [\langle NP, \mathtt{X}, W_0, W_1 \rangle], \epsilon \rangle \rightarrow$
$\quad \langle VP, \mathtt{X\hat{}Y}, W_1, W_2, [\langle NP, \mathtt{X}, W_0, W_1 \rangle], \epsilon \rangle \ \langle AdvP, \mathtt{Z}, W_2, W, \epsilon, \epsilon \rangle$

$\langle VP, \mathtt{X\hat{}mod(Y,Z)}, W_1, W, [\langle NP, \mathtt{X}, W_0, W_1 \rangle], \epsilon \rangle \rightarrow$
$\quad \langle VP, \mathtt{X\hat{}Y}, W_1, W_2, [\langle NP, \mathtt{X}, W_0, W_1 \rangle], \epsilon \rangle \ \langle PP, \mathtt{Z}, W_2, W, \epsilon, \epsilon \rangle$

$\langle S, \mathtt{sleep(X)}, W_0, W, \epsilon, \epsilon \rangle \rightarrow \langle NP, \mathtt{X}, W_0, [sleeps|W], \epsilon, \epsilon \rangle$

$\langle VP, \mathtt{X\hat{}sleep(X)}, [sleeps|W], W, [\langle NP, \mathtt{X}, W_0, [sleeps|W] \rangle], \epsilon \rangle \rightarrow$
$\quad \langle NP, \mathtt{X}, W_0, [sleeps|W], \epsilon, \epsilon \rangle$

$\langle S, \mathtt{see(X,Y)}, W_0, W, \epsilon, \epsilon \rangle \rightarrow \langle NP, \mathtt{X}, W_1, W, \epsilon, \epsilon \rangle \ \langle NP, \mathtt{Y}, W_0, [sees|W_1], \epsilon, \epsilon \rangle$

$\langle VP, \mathtt{Y\hat{}see(X,Y)}, [sees|W_0], W, [\langle NP, \mathtt{Y}, W_1, [sees|W_0] \rangle], \epsilon \rangle \rightarrow$
$\quad \langle NP, \mathtt{X}, W_0, W, \epsilon, \epsilon \rangle \ \langle NP, \mathtt{Y}, W_1, [sees|W_0], \epsilon, \epsilon \rangle$

$\langle PP, \mathtt{X\hat{}in(X)}, [in|W_0], W, \epsilon, \epsilon \rangle \rightarrow \langle NP, \mathtt{X}, W_0, W, \epsilon, \epsilon \rangle$

$\langle NP, \mathtt{john}, [John|W], W, \epsilon, \epsilon \rangle \rightarrow \epsilon$

$\langle NP, \mathtt{mary}, [Mary|W], W, \epsilon, \epsilon \rangle \rightarrow \epsilon$

$\langle NP, \mathtt{paris}, [Paris|W], W, \epsilon, \epsilon \rangle \rightarrow \epsilon$

$\langle AdvP, \mathtt{today}, [today|W], W, \epsilon, \epsilon \rangle \rightarrow \epsilon$

$\langle QM, \mathtt{ynq}, [?|W], W, \epsilon, \epsilon \rangle \rightarrow \epsilon$

Fig. 6. *Inverted sample grammar*

an argument position of the LHS logical form, rather than some position in
the input string.

New states are induced from old ones: For the indicated argument po-
sition, a possible logical form is selected and the dot is advanced one step
in all items where this particular logical form can occur in the current ar-
gument position, and the resulting new items constitute a new state. All
possible grammar symbols that can occur in the old argument position and
that can have this logical form are then collected. From these, all rules with
a matching LHS are invoked from the inverted grammar. Each such rule
will give rise to a new item where the dot marks the first argument position,
and the set of these new items will constitute another new state. If a new
set of items is constructed that is more specific than an existing one, then
this search branch is abandoned and the recursion terminates. If it on the other hand is more general, then it replaces the old one.

The state-construction phase starts off by creating an initial set consisting of a single dummy item with a dummy top grammar symbol and a dummy top logical form, corresponding to a dummy inverted grammar rule. In the sample grammar, this would be the rule $\langle S', \mathtt{f(X)}, W_0, W, \epsilon, \epsilon \rangle \rightarrow \langle S, \mathtt{X}, W_0, W, \epsilon, \epsilon \rangle$. The dot is at the beginning of the rule, selecting the first and only argument. The rest of the states are induced from this one. The first three states resulting from the inverted grammar of Figure 6 are shown in Figure 7, where the difference lists representing the word strings are omitted.

---

State 1
$\langle S', \mathtt{f(X)}, \epsilon, \epsilon \rangle \;\Rightarrow\; \bullet \langle S, \mathtt{X}, \epsilon, \epsilon \rangle$

State 2
$\langle S, \mathtt{mod(X,Y)}, \epsilon, \epsilon \rangle \;\Rightarrow\; \bullet \langle S, \mathtt{X}, \epsilon, \epsilon \rangle \; \langle QM, \mathtt{Y}, \epsilon, \epsilon \rangle$
$\langle S, \mathtt{mod(Y,Z)}, \epsilon, \epsilon \rangle \;\Rightarrow\; \bullet \langle VP, \mathtt{X\char`^Y}, [\langle NP, \mathtt{X} \rangle], \epsilon \rangle \; \langle AdvP, \mathtt{Z}, \epsilon, \epsilon \rangle$
$\langle S, \mathtt{mod(Y,Z)}, \epsilon, \epsilon \rangle \;\Rightarrow\; \bullet \langle VP, \mathtt{X\char`^Y}, [\langle NP, \mathtt{X} \rangle], \epsilon \rangle \; \langle PP, \mathtt{Z}, \epsilon, \epsilon \rangle$

State 3
$\langle S, \mathtt{mod(X,Y)}, \epsilon, \epsilon \rangle \;\Rightarrow\; \bullet \langle S, \mathtt{X}, \epsilon, \epsilon \rangle \; \langle QM, \mathtt{Y}, \epsilon, \epsilon \rangle$
$\langle S, \mathtt{mod(Y,Z)}, \epsilon, \epsilon \rangle \;\Rightarrow\; \bullet \langle VP, \mathtt{X\char`^Y}, [\langle NP, \mathtt{X} \rangle], \epsilon \rangle \; \langle AdvP, \mathtt{Z}, \epsilon, \epsilon \rangle$
$\langle S, \mathtt{mod(Y,Z)}, \epsilon, \epsilon \rangle \;\Rightarrow\; \bullet \langle VP, \mathtt{X\char`^Y}, [\langle NP, \mathtt{X} \rangle], \epsilon \rangle \; \langle PP, \mathtt{Z}, \epsilon, \epsilon \rangle$
$\langle VP, \mathtt{X\char`^mod(Y,Z)}, [\langle NP, \mathtt{X} \rangle], \epsilon \rangle \;\Rightarrow\; \bullet \langle VP, \mathtt{X\char`^Y}, [\langle NP, \mathtt{X} \rangle], \epsilon \rangle \; \langle AdvP, \mathtt{Z}, \epsilon, \epsilon \rangle$
$\langle VP, \mathtt{X\char`^mod(Y,Z)}, [\langle NP, \mathtt{X} \rangle], \epsilon \rangle \;\Rightarrow\; \bullet \langle VP, \mathtt{X\char`^Y}, [\langle NP, \mathtt{X} \rangle], \epsilon \rangle \; \langle PP, \mathtt{Z}, \epsilon, \epsilon \rangle$

Fig. 7. *The first three generation states*

---

The sets of items are used to compile the generation tables in the same way as is done for LR parsing. The goto entries correspond to transiting from one argument of a term to the next, and thus advancing the dot one step. The reductions correspond to applying the rules of items that have the dot at the end of the RHS, as is the case when LR parsing. There is no obvious analogy to the shift action — the closest thing would be the descend actions transiting from a functor to one of its arguments.

Note that there is no need to include the logical form of each lexicon entry in the generation tables. Instead, a typing of the logical forms can be introduced, and a representative of each type used in the actual tables, rather than the individual logical forms. This decreases the size of the tables drastically. For example, there is no point in distinguishing the states reached by traversing `john`, `mary` and `paris`, apart from ensuring that the correct word is added to the output word-string. This is accomplished much in the same way as preterminals, rather than individual words, figure in LR-parsing tables.

## 6   The Generation Algorithm

The generator works by recursive descent through the logical form while transiting between the internal states. It is driven by the descend, goto and reduce tables. A pushdown stack is used to store intermediate constituents. When generating a word string, the current state and logical form determine a transition to a new state, corresponding to the first argument of the logical form, through the *descend* table. A substring is generated recursively from the argument logical form, and this constituent is pushed onto the stack. The argument logical form, together with the new current state, determine a transition to the next state through the *goto* table. The next state corresponds to the next argument of the original logical form, and another substring is generated from this argument logical form, etc. When no more arguments remain, an inverted grammar rule is selected nondeterministically by the *reduce* table and applied to the top portion of the stack, constructing a word string corresponding to the original logical form and completing this generation cycle.[4]

We now turn to optimizing the generation tables.

## 7   Optimizing the Generation Tables

The basic idea underlying the optimization technique presented in this article is to remove as much nondeterminism from the generation tables as possible. One problem is that it may be impossible to remove all nondeterminism for the simple reason that the current piece of logical form may in fact allow multiple paraphrases. In this case, we say that we have "real" nondeterminism. On the other hand, it may be the case that although locally, several alternatives are possible, subsequent generation may rule out all but one of them. We will call this "spurious" nondeterminism.

---

[4]This is a bottom-up rule invocation scheme. It could easily be modified so that a rule is instead applied *before* constructing the substrings recursively, resulting in a top-down rule-invocation scheme.

Due to the grammar inversion, and the way the sets of items are constructed, all LHS logical forms of the items in some particular state will be the same, and will thus have equal arity. Thus, there will be nothing analogous to shift-reduce conflicts in the resulting generation tables, only reduce-reduce conflicts. This means that the latter is the sole source of nondeterminism, and that this will arise only in states with more than one possible reduction. By inspecting the number of items left in each "reductive state", i.e., each state where the dot is at the end of the rules, we can determine whether or not the generation tables will be deterministic.

The logical form can be inspected down to an arbitrary depth of recursion when compiling the sets of items, and this parameter can be varied. This is closely related to the use of lookahead symbols in an LR parser; increasing the depth is analogous to increasing the number of lookahead symbols. The amount of semantic lookahead will be reflected in the goto and descend table entries. No semantic lookahead would mean only taking the functor of the logical form into consideration, and in the example above, a typical action table entry would be `descend(1,mod(_,_),2)`.[5] This would mean that the generator would operate on State 2 of Figure 7 when generating from the first argument of the `mod/2` term, and both the $S$ alternative and the (merged) $VP$ alternative(s) would be attempted nondeterministically.

By taking the arguments of the logical form into account, the degree of nondeterminism can be reduced, and for the grammar given in Figure 1, it is eliminated completely. In the example, if the second argument of the `mod/2` term is `ynq`, then only the $S$ alternative will be considered when generating from the first argument, since the relevant descend entries and states will be those of Figure 8. The optimal depth may vary for each individual table entry, and even within it, and a scheme has been devised to automatically find such an optimum.

Assuming that it is actually possible to construct fully deterministic generation tables by filtering on a large enough amount of semantic lookahead, the problem reduces to for each table entry finding a lookahead depth that will result in only one single remaining item in each reductive state. This is in fact a stronger requirement than that all nondeterminism be spurious: It may be the case that for each possible logical form, it is possible to determine the appropriate reduction by a sufficient amount of semantic lookahead, but due to potentially infinite recursion, no preassigned limit on

---

[5]Here "_" denotes a don't-care variable.

```
descend(1, mod(mod(_,_),ynq), 2A).
descend(1, mod(see(_,_),ynq), 2B).
descend(1, mod(sleep(_),ynq), 2C).
```

State 2A
$\langle S, \texttt{mod(mod(X,Y),ynq)}, \epsilon, \epsilon \rangle \;\; \Rightarrow \;\; \bullet \, \langle S, \texttt{mod(X,Y)}, \epsilon, \epsilon \rangle \, \langle QM, \texttt{ynq}, \epsilon, \epsilon \rangle$

State 2B
$\langle S, \texttt{mod(see(X,Y),ynq)}, \epsilon, \epsilon \rangle \;\; \Rightarrow \;\; \bullet \, \langle S, \texttt{see(X,Y)}, \epsilon, \epsilon \rangle \, \langle QM, \texttt{ynq}, \epsilon, \epsilon \rangle$

State 2C
$\langle S, \texttt{mod(sleep(X),ynq)}, \epsilon, \epsilon \rangle \;\; \Rightarrow \;\; \bullet \, \langle S, \texttt{sleep(X)}, \epsilon, \epsilon \rangle \, \langle QM, \texttt{ynq}, \epsilon, \epsilon \rangle$

Fig. 8. *Alternative generation states*

it will do. This is elaborated in the following section.

The scheme employs iterative deepening. It tries to construct fully deterministic tables by first allowing a total amount of semantic lookahead of one, then of two, etc., up to some maximum limit. This is however not done globally, but at each recursive call to the sets-of-items construction step, when a piece of logical form and a set of grammar symbols are used to invoke new inverted grammar rules to construct new sets of items.

At this point, the total amount of available lookahead is distributed through the arguments of the functor of the current piece of logical form, and then further down to the arguments of the arguments, etc., until all has been used up. The current sets of items are then tentatively constructed. Increased semantic-lookahead depth will split potential nondeterminism in the resulting reductive states into distinct sets of items, and thus into distinct reductive states with less nondeterminism, or preferably, with no nondeterminism at all. If the resulting reductive states are all deterministic, then this particular semantic-lookahead setting is used to compile the actual generation tables, and the scheme recurses. In more detail, a set of terms mirroring the various ways of assigning semantic lookahead are generated and ordered according to how much lookahead they use up. The first one to yield fully deterministic reductive states is used when constructing the actual tables and is passed down in the recursion.

In the running example, the first argument of `mod/2` contributes no important information when descending from State 1, while the second one does. The scheme correctly finds the optimal depths when transiting from State 1, resulting in the State 2 and descend entry of Figure 9.

---

$$\texttt{descend(1, mod(\_,ynq), 2)}.$$

State 2
$$\langle S, \texttt{mod(X,ynq)}, \epsilon, \epsilon \rangle \;\Rightarrow\; \bullet\, \langle S, \texttt{X}, \epsilon, \epsilon \rangle \; \langle QM, \texttt{ynq}, \epsilon, \epsilon \rangle$$

Fig. 9. *Alternative alternative generation states*

---

Since the scheme employs iterative deepening, this will guarantee that locally, no alternative table entries can inspect a smaller portion of the logical forms and still be deterministic, given the previous choices of semantic lookahead. This is a greedy algorithm, and it could potentially be the case that another choice of semantic lookahead would lead to less required lookahead in total by reducing that of the table entries generated in later recursion steps.

## 8  An Example-Based Optimization Technique

The optimization scheme as described so far is limited to grammars without real nondeterminism that only have removable spurious nondeterminism. A simple way of extending this to more general grammars is to introduce a second outer level of iterative deepening controlling the amount of nondeterminism tolerated in each recursive call to the sets-of-items construction step. First, we try to construct generation tables with only one reduction in each reductive state. If this proves impossible within the maximum amount of total semantic lookahead allowed, we try to construct tables with at most two reductions in each resulting reductive state, etc. Since there is a finite number of inverted grammar rules, and thus a finite number of possible items, this process will terminate. Again, this optimization is done locally at each recursive call to the sets-of-items construction step.

A problem with this approach is that the number of possible ways of assigning semantic lookahead increases drastically with the amount of lookahead allowed, and some heuristics are needed to direct the search. We will shortly describe a method that constructs more fine-tuned generation tables by using training examples to guide the search; to determine how much real nondeterminism there is at each point that cannot be removed; and to find appropriate lookahead depths that will remove all spurious nondeterminism on the training corpus. First, we will however examine spurious nondeterminism a bit closer.

Assume that we add the following grammar rules for handling *NP*s with internal structure:[6]

$$\langle NP, \mathtt{q(X,Y)} \rangle \;\rightarrow\; \langle Det, \mathtt{X} \rangle \; \langle \bar{N}, \mathtt{Y} \rangle$$
$$\langle \bar{N}, \mathtt{X} \rangle \;\rightarrow\; \langle N, \mathtt{X} \rangle$$
$$\langle \bar{N}, \mathtt{mod(X,Y)} \rangle \;\rightarrow\; \langle \bar{N}, \mathtt{Y} \rangle \; \langle \bar{N}, \mathtt{X} \rangle$$
$$\langle \bar{N}, \mathtt{mod(X,Y)} \rangle \;\rightarrow\; \langle AP, \mathtt{Y} \rangle \; \langle N, \mathtt{X} \rangle$$
$$\langle AP, \mathtt{mod(X,Y)} \rangle \;\rightarrow\; \langle N, \mathtt{Y} \rangle \; \langle AP, \mathtt{X} \rangle$$
$$\langle AP, \mathtt{X} \rangle \;\rightarrow\; \langle A, \mathtt{X} \rangle$$

This will allow derivations like that of Figure 10. Here `APoNB` reads "Adjective phrase or N-bar". This in turn will allow constructing logical forms like

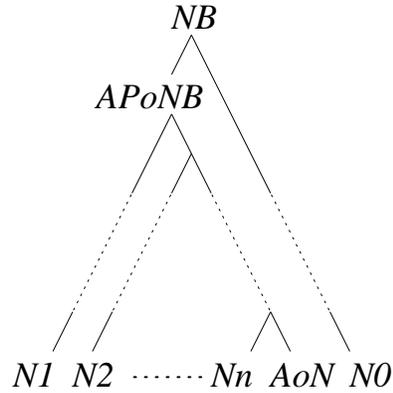

Fig. 10. *A sample derivation*

`mod(N0,mod(mod(...mod(AoN,Nn),...,N2),N1))`. To determine which of the rules $\langle \bar{N}, \mathtt{mod(X,Y)} \rangle \rightarrow \langle \bar{N}, \mathtt{Y} \rangle \langle \bar{N}, \mathtt{X} \rangle$ and $\langle \bar{N}, \mathtt{mod(X,Y)} \rangle \rightarrow \langle AP, \mathtt{Y} \rangle \langle N, \mathtt{X} \rangle$ to apply, we must inspect the first argument `AoN` — verb or noun — of the innermost `mod/2` term, which may be arbitrarily deeply nested. Although this will never introduce multiple paraphrases, it does allow spurious nondeterminism that cannot be handled by a bounded amount of semantic lookahead.

A highly respectable objection to the presented example is that, apart from the proposed treatment of noun-noun and noun-adjective compounds being linguistically somewhat dubious, we will in practice never see cases where we need a very large amount of semantic lookahead. Precisely this is one of the two corner stones on which the example-based optimization technique presented in this section rests. The other one is the observation that a lower bound of the amount of real nondeterminism can easily be established for each (portion of a) training example, while it is in the general case difficult to do this directly from the grammar.

---

[6] Again, the difference lists representing the word strings have been omitted

Thus, the training examples are used for three purposes: Firstly, to limit the search to search branches that are relevant for input data that actually occur in real life. Secondly, to establish the minimum amount of nondeterminism at each point, i.e., the amount of real nondeterminism at this point that cannot be removed by greater lookahead depth. Thirdly, to find appropriate lookahead depths that will remove all spurious nondeterminism at each point in the training example.

The generation tables are constructed much in the same way as in the previous section. The main difference is that instead of aiming at full determinism, the target nondeterminism is the real nondeterminism at each point of each training example. In more detail, a set of terms mirroring the various ways of assigning semantic lookahead are generated from the set of training examples, and they are ordered according to how much lookahead they employ. Intuitively, a (sub)term is constructed from each training example by replacing parts of it with free variables, thus removing the information contained in these parts of the training example, and the subterms are merged to form one term. Thus, terms employing more lookahead will contain more detailed information from the set of training examples. The first term to yield as deterministic reductive states as the one corresponding to the set of whole training examples, where no information has been blocked out by variables, is used for constructing the actual tables and is passed down in the recursion.

A technical complication is that the training examples interact with the termination criteria of the sets-of-items construction step: Although a new set of items may be more specific than an old one, it may stem from more demanding training examples. In the current version of the scheme, this would result in recompiling the sets of items from the earliest point where too simple examples were used, this time including the more demanding examples. To handle input outside the training corpus, a default lookahead depth is assigned to the possible continuations that are not encountered among the training examples. This means that the resulting generation tables preserve completeness and are guaranteed to be optimal, modulo the limitations of greedy algorithms, for input sufficiently similar to combinations of examples in the training corpus, but not necessarily for other input.

The degree of generalization is considerable: To return to the running example of the nondeterminism in State 2 discussed above, a single training example like (a logical form corresponding to) *John sleeps?* or *Mary sees a house in Paris* will remove all nondeterminism in this state. In general, the table size seems to increase moderately with the number of training examples due to the good degree of generalization, although this needs to be more thoroughly investigated.

The modified algorithm for including the training examples into the LR-compilation algorithm is guaranteed to terminate if the original LR-compilation algorithm terminates. The worst-case complexity is however not very good. However, for the grammars and training sets tested this far, processing efficiency is not a problem, though we can envision that for considerably larger grammars and training sets, there will be a need for optimizing the optimization procedure further.

## 9   Discussion

The new generation algorithm constitutes an improvement on the semantic-head-driven generation algorithm that allows "functor merging", i.e., enables processing various grammar rules, or rule combinations, that introduce the same semantic structure simultaneously, thereby greatly reducing the search space. The algorithm proceeds by recursive descent through the logical form, and using the terminology of the SHDG algorithm, what the new algorithm in effect does is to process all chains from a particular set of grammar symbols down to some particular piece of logical form in parallel until a reduction is attempted, rather than to construct and try each one separately in turn. This requires a grammar-inversion technique that is fundamentally different from techniques such as the essential-argument algorithm, see the following, since it must display the grammar from the point of view of the logical form, rather than from that of the word string. LR-compilation techniques accomplish the functor merging by compiling the inverted grammar into a set of generation tables.

The grammar inversion rearranges the grammar as a whole according to the functor-argument structure of the logical forms. Other inversion schemes, such as the essential-argument algorithm (Strzalkowski 1990) or the direct-inversion approach (Minnen et al. 1995), are mainly concerned with locally rearranging the order of the RHS constituents of individual grammar rules by examining the flow of information through these constituents, to ensure termination and increase efficiency. Although this can occasionally change the set of RHS symbols in a rule, it is done to these ends, rather than to reflect the functor-argument structure.

Although the sample grammar used throughout the article is essentially context-free, there is nothing in principle that restricts the method to such grammars. In fact, the method could be extended to grammars employing complex feature structures as easily as the LR-parsing scheme itself, see for example (Nakazawa 1991), and this is currently being done. Some hand editing is necessary when preparing the grammar for the inversion step, but it is limited to specifying the flow of arguments in the grammar rules. Furthermore, this could potentially be fully automated.

The set of applicable reductions can be diminished by resorting to deeper semantic lookahead, at the price of a larger number of internal states, and there is in general a tradeoff between the size of the resulting generation tables and the amount of nondeterminism when reducing. The employed amount of semantic lookahead can be varied, and a scheme has been devised and tested that automatically determines appropriate tradeoff points, optionally based on a collection of training examples.

The latter version of the scheme turns out to be related to explanation-based learning (EBL) which has proved quite successful for optimizing *LR-parsing* tables for syntactic analysis. There, the basic idea is to learn special grammar rules from the original ones and a set of training examples by chunking together the former based on how they are used to parse the latter. The relevant references are (Samuelsson & Rayner 1991), (Samuelsson 1994a) and (Neumann 1994).

Rayner and Samuelsson basically trade coverage for speed and accuracy by using the training examples to compile a new grammar that is used instead of the original one. Their problem is that the underlying NL systems that they work on employ find-all parsing strategies and subsequent selection of the preferred analysis. This makes it very difficult to integrate the learned grammar with the original one without losing all processing speed gained. Neumann strives for a very close integration between the learned and original grammars by falling back to the original grammar when processing with the learned grammar alone has proved insufficient. He utilizes the fact that his original system employs a best-first parsing strategy, which allows intelligent reuse of partial results from the attempt to parse with the learned grammar.

Another problem that has not previously been satisfactorily resolved is how to determine the degree of generalization of the examples, or viewed from another point of view, how to chunk together the original grammar rules. Rayner and Neumann hand-code special meta-rules, so-called operationality criteria, for this based on linguistic intuition. These criteria are then refined manually by experimentation. Samuelsson offers an automatic method for doing this that relates the desired coverage to the way the examples are generalized (Samuelsson 1994b). This quantity is however only indirectly related to the actual performance of the system using the resulting learned grammar.

In contrast to this, the method described in the current article automatically preserves completeness; achieves fully seamless integration, since there is only one processing mode; and automatically determines the degree of generalization by minimizing a quantity that has a profound direct influence on the resulting performance, namely the amount of nondeterminism in each reductive state. It would be very interesting to see if this idea could be carried over to syntactic parsing by manipulating the number of lookahead symbols to minimize the number of shift-reduce and reduce-reduce conflicts in the resulting LR parsing tables.

The method has been implemented and applied to more complex grammars than the simple one used as an example in this article, and it works excellently. Although these grammars are still too naive to form the basis of a serious empirical evaluation lending substantial experimental support to the method as a whole, it should be obvious from the algorithm itself that the reduction in search space compared to the SHDG algorithm is most substantial. Nonetheless, such an evaluation is a top-priority item on the future-work agenda.

# Index